\documentclass[fleqn,10pt]{olplainarticle}
\usepackage{siunitx}
\usepackage{cite}
\usepackage{amsmath,amssymb,amsfonts}
\usepackage{mathtools}
\usepackage{algorithmic}
\usepackage{graphicx}
\usepackage{textcomp}
\usepackage{xcolor}
\usepackage{multirow}
\usepackage{comment}
\usepackage{csquotes}
\usepackage[utf8]{inputenc}
\usepackage{hyperref}
\usepackage{blindtext}

\DeclareUnicodeCharacter{FB01}{fi}
\DeclareUnicodeCharacter{FFFC}{}
\newcommand*\samethanks[1][\value{footnote}]{\footnotemark[#1]}

\usepackage{lipsum}

\newcommand\blfootnote[1]{%
	\begingroup
	\renewcommand\thefootnote{}\footnote{#1}%
	\addtocounter{footnote}{-1}%
	\endgroup
}

\title{A Survey and Taxonomy of Adversarial Neural Networks for Text-to-Image Synthesis\blfootnote{Wiley Interdisciplinary Reviews: Data Mining and Knowledge Discovery. 2019}}

\author{\mdseries{Jorge Agnese$^{1}$\footnote{Equally contributing authors.} , Jonathan Herrera$^{1}$\samethanks[1] , Haicheng Tao$^{2}$\samethanks[1] , Xingquan Zhu$^{1}$}\\
	\mdseries{$^{1}$Department of Computer \& Electrical Engineering and Computer Science, \protect\\  Florida Atlantic University, Boca Raton, FL 33431, USA \protect\\  Email: jagnese2018@fau.edu, herreraj2015@fau.edu, xzhu3@fau.edu}\\
	\mdseries{$^{2}$Jiangsu Provincial Key Laboratory of E-Business, \protect\\Nanjing University of Finance and Economics, Nanjing, China, \protect\\ Email: tao005@mail.ustc.edu.cn}\\
}

\keywords{Text-to-image synthesis, generative adversarial network (GAN), deep learning, machine learning}

\begin{abstract}

Text-to-image synthesis refers to computational methods which translate human written textual descriptions, in the form of keywords or sentences, into images with similar semantic meaning to the text. In earlier research, image synthesis relied mainly on word to image correlation analysis combined with supervised methods to find best alignment of the visual content matching to the text. Recent progress in deep learning (DL) has brought a new set of unsupervised deep learning methods, particularly deep generative models which are able to generate realistic visual images using suitably trained neural network models. The change of direction from the computer vision based approaches to artificial intelligence (AI) driven methods ignited the intense interest in industry, such as virtual reality, recreational \& professional (eSports) gaming, and computer-aided design \textit{etc.}, to automatically generate compelling images from text-based natural language descriptions.

 In this paper, we review the most recent development in the text-to-image synthesis research domain. Our goal is to provide value by delivering a comparative review of the state-of-the-art models in terms of their architecture and design. Our survey first introduces image synthesis and its challenges, and then reviews key concepts such as generative adversarial networks (GANs) and deep convolutional encoder-decoder neural networks (DCNN). After that, we propose a taxonomy to summarize GAN based text-to-image synthesis into four major categories: Semantic Enhancement GANs, Resolution Enhancement GANs, Diversity Enhancement GANS, and Motion Enhancement GANs. We elaborate the main objective of each group, and further review typical GAN architectures in each group. The taxonomy and the review outline the techniques and the evolution of different approaches, and eventually provide a clear roadmap to summarize the list of contemporaneous solutions that utilize GANs and DCNNs to generate enthralling results in categories such as human faces, birds, flowers, room interiors, object reconstruction from edge maps (games) etc. The survey will conclude with a comparison of the proposed solutions, challenges that remain unresolved, and future developments in the text-to-image synthesis domain.
\end{abstract}

\begin{document}
	\begin{sloppypar}
\maketitle

\section{Introduction}

\begin{displayquote}
	`` (GANs), and the variations that are now being proposed is the most interesting idea in the last 10 years in ML, in my opinion.'' (2016)
	\begin{flushright}-- Yann LeCun  \end{flushright}
\end{displayquote}

A picture is worth a thousand words! While written text provide efficient, effective, and concise ways for communication, visual content, such as images, is a more comprehensive, accurate, and intelligible method of information sharing and understanding.  Generation of images from text descriptions, \textit{i.e.} text-to-image synthesis, is a complex computer vision and machine learning problem that has seen great progress over recent years. Automatic image generation from natural language may allow users to describe visual elements through visually-rich text descriptions. The ability to do so effectively is highly desirable as it could be used in artificial intelligence applications such as computer-aided design, image editing \citep{chen:cvpr:2018,Yan:TOG:2016}, game engines for the development of the next generation of video games\citep{fcn}, and pictorial art generation \citep{Elgammal:can:2017}.

\subsection{\textcolor{black}{Traditional Learning Based Text-to-image Synthesis}}
\begin{figure}
	\centering
	\includegraphics[width=0.7\columnwidth]{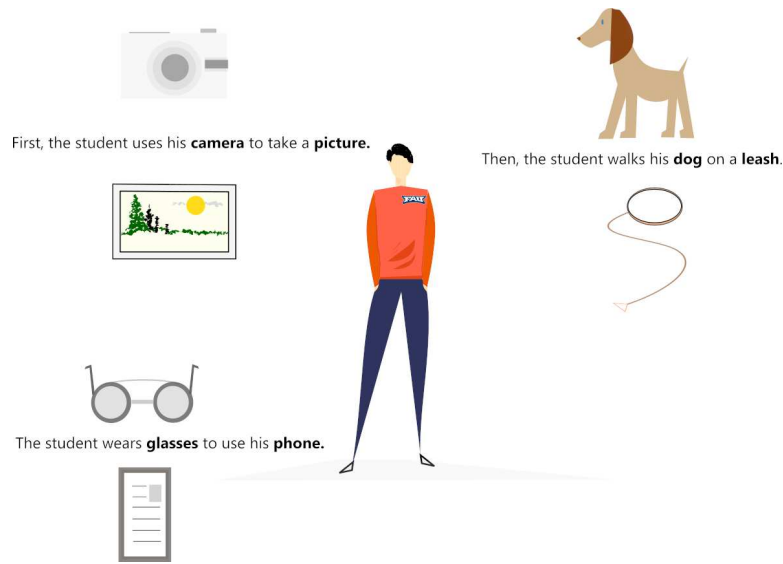}
	\caption{Early research on text-to-image synthesis~\citep{zhu:AAAI:2007}. The system uses correlation between keywords (or keyphrase) and images and identifies informative and ``picturable'' text units, then searches for the most likely image parts conditioned on the text, and eventually optimizes the picture layout conditioned on both the text and image parts. }\label{fig.early.learning}
\end{figure}

In the early stages of research, text-to-image synthesis was mainly carried out through a search and supervised learning combined process~\citep{zhu:AAAI:2007}, as shown in Figure~\ref{fig.early.learning}. In order to connect text descriptions to images, one could use correlation between keywords (or keyphrase) \& images that identifies informative and ``picturable'' text units; then, these units would search for the most likely image parts conditioned on the text, eventually optimizing the picture layout conditioned on both the text and the image parts. Such methods often integrated multiple artificial intelligence key components, including natural language processing, computer vision, computer graphics, and machine learning.

\begin{figure}
	\centering
	\includegraphics[width=0.7\columnwidth]{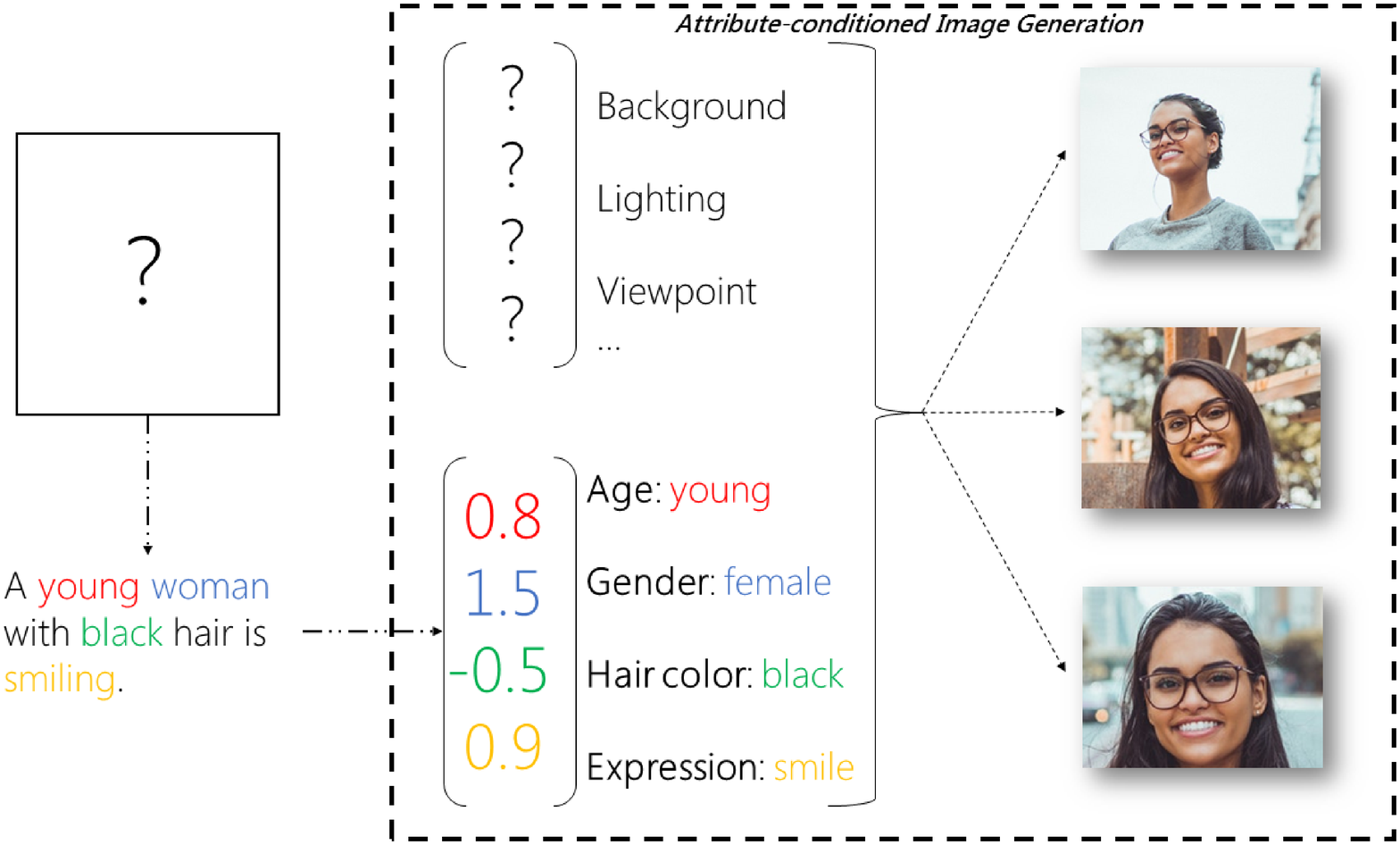}
	\caption{Supervised learning based text-to-image synthesis~\citep{Yan:Attribute2Image:2016}. The supervised learning process aims to learn layered generative models to generate visual content. Because the learning is customized/conditioned by the given attributes, the generative models of Attribute2Image can generative images with respect to different attributes, such as hair color, age, \textit{etc.}} \label{fig.supervised.learning}
\end{figure}

The major limitation of the traditional learning based text-to-image synthesis approaches is that they lack the ability to generate new image content; they can only change the characteristics of the given/training images. Alternatively, research in generative models has advanced significantly and delivers solutions to learn from training images and produce new visual content. For example, Attribute2Image~\citep{Yan:Attribute2Image:2016} models each image as a composite of foreground and background. In addition, a layered generative model with disentangled latent variables is learned, using a variational auto-encoder, to generate visual content. Because the learning is customized/conditioned by given attributes, the generative models of Attribute2Image can generate images with respect to different attributes, such as gender, hair color, age, \textit{etc.}, as shown in Figure~\ref{fig.supervised.learning}.

\begin{figure}
	\centering
	\includegraphics[width=\columnwidth]{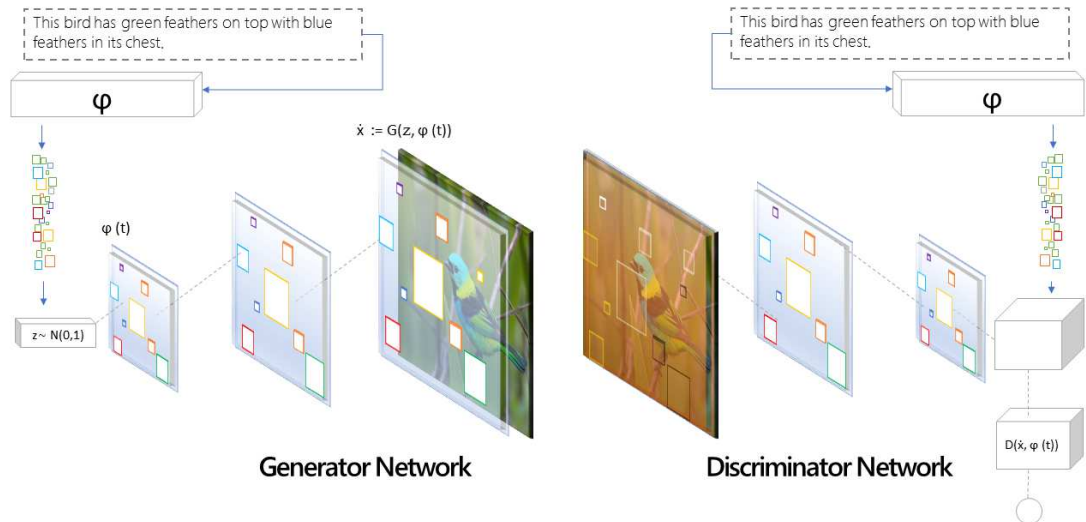}
	\caption{Generative adversarial neural network (GAN) based text-to-image synthesis~\citep{sur2}. GAN based text-to-image synthesis combines discriminative and generative learning to train neural networks resulting in the generated images semantically resemble to the training samples or tailored to a subset of training images (\textit{i.e.} conditioned outputs). $\varphi()$ is a feature embedding function, which converts text as feature vector. $z$ is a latent vector following normal distributions with zero mean.  $\hat{x}=G(z,\varphi(t)$ denotes a synthetic image generated from the generator, using latent vector $z$ and the text features $\varphi(t)$ as the input. $D(\hat{x},\varphi(t))$ denotes the prediction of the discriminator based on the input $\hat{x}$ the generated image and $\varphi(t)$ text information of the generated image. The explanations about the generators and discriminators are detailed in Section \ref{sec.gan}.}
	\label{fig.gan.t2i}
\end{figure}

\subsection{GAN Based Text-to-image Synthesis}
Although generative model based text-to-image synthesis provides much more realistic image synthesis results, the image generation is still conditioned by the limited attributes. In recent years, several papers have been published on the subject of text-to-image synthesis. Most of the contributions from these papers rely on multimodal learning approaches that include generative adversarial networks and deep convolutional decoder networks as their main drivers to generate entrancing images from text \citep{sur1,reed1,gan,attn,aux}. \par

First introduced by Ian Goodfellow et al. \citep{gan}, generative adversarial networks (GANs) consist of two neural networks paired with a discriminator and a generator. These two models compete with one another, with the generator attempting to produce synthetic/fake samples that will fool the discriminator and the discriminator attempting to differentiate between real (genuine) and synthetic samples. Because GANs' adversarial training aims to cause generators to produce images similar to the real (training) images, GANs can naturally be used to generate synthetic images (image synthesis), and this process can even be customized further by using text descriptions to specify the types of images to generate, as shown in Figure~\ref{fig.gan.t2i}. \par

\begin{figure}
	\centering
	\includegraphics[width=\columnwidth]{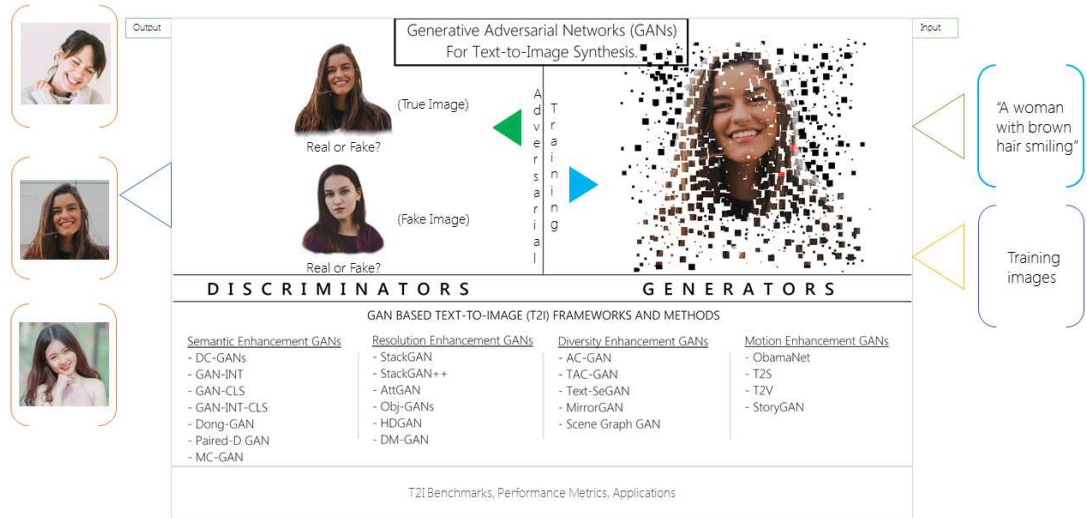}
	\caption{A visual summary of GAN based text-to-image (T2I) synthesis process, and the summary of GAN based frameworks/methods reviewed in the survey. }\label{fig.visualabstract}
\end{figure}

Much like text-to-speech and speech-to-text conversion, there exists a wide variety of problems that text-to-image synthesis could solve in the computer vision field specifically \citep{reed1, s2t}. Nowadays, researchers are attempting to solve a plethora of computer vision problems with the aid of deep convolutional networks, generative adversarial networks, and a combination of multiple methods, often called multimodal learning methods \citep{reed1}. For simplicity, \textit{multiple learning methods} will be referred to as \textit{multimodal learning} hereafter \citep{mdl}. Researchers often describe multimodal learning as a method that incorporates characteristics from several methods, algorithms, and ideas. This can include ideas from two or more learning approaches in order to create a robust implementation to solve an uncommon problem or improve a solution \citep{reed1,lrgan,fastconv,tacgan,mmlsur}. \par

\textcolor{black}
{
	In this survey, we focus primarily on reviewing recent works that aim to solve the challenge of text-to-image synthesis using generative adversarial networks (GANs). In order to provide a clear roadmap, we propose a taxonomy to summarize reviewed GANs into four major categories. Our review will elaborate the motivations of methods in each category, analyze typical models, their network architectures, and possible drawbacks for further improvement. The visual abstract of the survey and the list of reviewed GAN frameworks is shown in Figure \ref{fig.visualabstract}.
}

\textcolor{black}
{
	The remainder of the survey is organized as follows. Section 2 presents a brief summary of existing works on subjects similar to that of this paper and highlights the key distinctions making ours unique. Section 3 gives a short introduction to GANs and some preliminary concepts related to image generation, as they are the engines that make text-to-image synthesis possible and are essential building blocks to achieve photo-realistic images from text descriptions. Section 4 proposes a taxonomy to summarize GAN based text-to-image synthesis, discusses models and architectures of novel works focused solely on text-to-image synthesis. This section will also draw key contributions from these works in relation to their applications. Section 5 reviews GAN based text-to-image synthesis benchmarks, performance metrics, and comparisons, including a simple review of GANs for other applications. In section 6, we conclude with a brief summary and outline ideas for future interesting developments in the field of text-to-image synthesis.} \par

\section{Related Work}
With the growth and success of GANs, deep convolutional decoder networks, and multimodal learning methods, these techniques were some of the first procedures which aimed to solve the challenge of image synthesis. Many engineers and scientists in computer vision and AI have contributed through extensive studies and experiments, with numerous proposals and publications detailing their contributions. Because GANs, introduced by \citet{gan}, are emerging research topics, their practical applications to image synthesis are still in their infancy. Recently, many new GAN architectures and designs have been proposed to use GANs for different applications, \textit{e.g.} using GANs to generate sentimental texts~\citep{wang:ijcai:2018}, or using GANs to transform natural images into cartoons~\citep{chen:cvpr:2018:cargoon}. \par

Although GANs are becoming increasingly popular, very few survey papers currently exist to summarize and outline contemporaneous technical innovations and contributions of different GAN architectures~\citep{hong:CSUR:2019,GAN_OV}. Survey papers specifically attuned to analyzing different contributions to text-to-image synthesis using GANs are even more scarce. We have thus found two surveys \citep{sur2,sur1} on image synthesis using GANs, which are the two most closely related publications to our survey objective. In the following paragraphs, we briefly summarize each of these surveys and point out how our objectives differ from theirs. \par

In \citet{sur2}, the authors provide an overview of image synthesis using GANs. In this survey, the authors discuss the motivations for research on image synthesis and introduce some background information on the history of GANs, including a section dedicated to core concepts of GANs, namely generators, discriminators, and the min-max game analogy, and some enhancements to the original GAN model, such as conditional GANs, addition of variational auto-encoders, \textit{etc.}. In this survey, we will carry out a similar review of the background knowledge because the understanding of these preliminary concepts is paramount for the rest of the paper. Three types of approaches for image generation are reviewed, including direct methods (single generator and discriminator), hierarchical methods (two or more generator-discriminator pairs, each with a different goal), and iterative methods (each generator-discriminator pair generates a gradually higher-resolution image). Following the introduction, \citet{sur2} discusses methods for text-to-image and image-to-image synthesis, respectively, and also describes several evaluation metrics for synthetic images, including inception scores and Frechet Inception Distance (FID), and explains the significance of the discriminators acting as learned loss functions as opposed to fixed loss functions.

Different from the above survey, which has a relatively broad scope in GANs, our objective is heavily focused on text-to-image synthesis. Although this topic, text-to-image synthesis, has indeed been covered in \citet{sur2}, they did so in a much less detailed fashion, mostly listing the many different works in a time-sequential order. In comparison, we will review several representative methods in the field and outline their models and contributions in detail. \par

Similarly to \citet{sur2}, the second survey paper \citep{sur1} begins with a standard introduction addressing the motivation of image synthesis and the challenges it presents followed by a section dedicated to core concepts of GANs and enhancements to the original GAN model. In addition, the paper covers the review of two types of applications: (1) unconstrained applications of image synthesis such as super-resolution, image inpainting, \textit{etc.}, and (2) constrained image synthesis applications, namely image-to-image, text-to-image, and sketch-to image, and also discusses image and video editing using GANs. Again, the scope of this paper is intrinsically comprehensive, while we focus specifically on text-to-image and go into more detail regarding the contributions of novel state-of-the-art models. \par

Other surveys have been published on related matters, mainly related to the advancements and applications of GANs \citep{sur3, sur4}, but we have not found any prior works which focus specifically on text-to-image synthesis using GANs. To our knowledge, this is the first paper to do so.

\textcolor{black}{
	\section{Preliminaries and Frameworks}
	In this section, we first introduce preliminary knowledge of GANs and one of its commonly used variants, conditional GAN (\textit{i.e.} cGAN), which is the building block for many GAN based text-to-image synthesis models. After that, we briefly separate GAN based text-to-image synthesis into two types, Simple GAN frameworks \textit{vs.} Advanced GAN frameworks, and discuss why advanced GAN architecture for image synthesis.}

\textcolor{black}
{
	Notice that the simple \textit{vs.} advanced GAN framework separation is rather too brief, our taxonomy in the next section will propose a taxonomy to summarize advanced GAN frameworks into four categories, based on their objective and designs.
}

\subsection{Generative Adversarial Neural Network}\label{sec.gan}
Before moving on to a discussion and analysis of works applying GANs for text-to-image synthesis, there are some preliminary concepts, enhancements of GANs, datasets, and evaluation metrics that are present in some of the works described in the next section and are thus worth introducing. \par
\begin{figure}
	\includegraphics[width=\columnwidth]{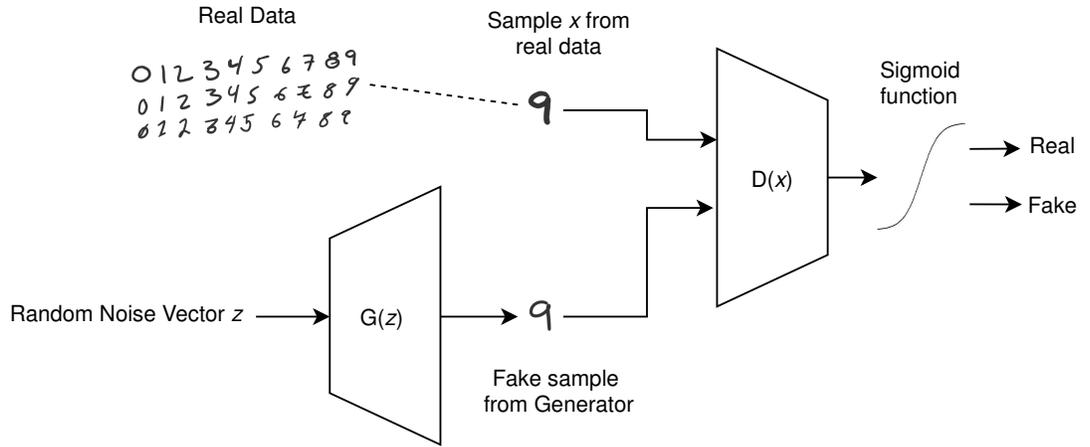}
	\caption{A conceptual view of the Generative Adversarial Network (GAN) architecture. The Generator G(z) is trained to generate synthetic/fake resemble to real samples, from a random noise distribution. The fake samples are fed to the Discriminator D(x) along with real samples. The Discriminator is trained to differentiate fake samples from real samples. The iterative training of the generator and the discriminator helps GAN deliver good generator generating samples very close to the underlying training samples.} \label{fig.GAN.structure}
\end{figure}

As stated previously, GANs were introduced by Ian Goodfellow et al. \citep{gan} in 2014, and consist of two deep neural networks, a generator and a discriminator, which are trained independently with conflicting goals: The generator aims to generate samples closely related to the original data distribution and fool the discriminator, while the discriminator aims to distinguish between samples from the generator model and samples from the true data distribution by calculating the probability of the sample coming from either source. A conceptual view of the generative adversarial network (GAN) architecture is shown in Figure~\ref{fig.GAN.structure}.

The training of GANs is an iterative process that, with each iteration, updates the generator and the discriminator with the goal of each defeating the other. leading each model to become increasingly adept at its specific task until a threshold is reached. This is analogous to a min-max game between the two models, according to the following equation:
\begin{equation}
\min_{\theta_g} \max_{\theta_d} V(D_{\theta_d}, G_{\theta_g}) = \mathbb{E}_{x \sim Pdata(x)}[\log(D_{\theta_d}(x))] +
\mathbb{E}_{x \sim P_z(z)}[\log(1-D_{\theta_d}(G_{\theta_g}(z)))]
\label{eq.gan.obj}
\end{equation}
In Eq.~(\ref{eq.gan.obj}), $x$ denotes a multi-dimensional sample, \textit{e.g.,} an image, and $z$ denotes a multi-dimensional latent space vector, \textit{e.g.,} a multidimensional data point following a predefined distribution function such as that of normal distributions. $D_{\theta_d}()$ denotes a discriminator function, controlled by parameters $\theta_d$, which aims to classify a sample into a binary space. $G_{\theta_g}()$ denotes a generator function, controlled by parameters $\theta_g$, which aims to generate a sample from some latent space vector. For example, $G_{\theta_g}(z)$ means using a latent vector $z$ to generate a synthetic/fake image, and $D_{\theta_d}(x)$ means to classify an image $x$ as binary output (\textit{i.e.} true/false or 1/0). In the GAN setting, the discriminator $D_{\theta_d}()$ is learned to distinguish a genuine/true image (labeled as 1) from fake images (labeled as 0). Therefore, given a true image $x$, the ideal output from the discriminator $D_{\theta_d}(x)$ would be 1. Given a fake image generated from the generator $G_{\theta_g}(z)$, the ideal prediction from the discriminator $D_{\theta_d}(G_{\theta_g}(z))$ would be 0, indicating the sample is a fake image.

Following the above definition, the $\min\max$ objective function in Eq.~(\ref{eq.gan.obj}) aims to learn parameters for the discriminator ($\theta_d$) and generator ($\theta_g$) to reach an optimization goal: The discriminator intends to differentiate true \textit{vs.} fake images with maximum capability $\max_{\theta_d}$ whereas the generator intends to minimize the difference between a fake image \textit{vs.} a true image $\min_{\theta_g}$. In other words, the discriminator sets the characteristics and the generator produces elements, often images, iteratively until it meets the attributes set forth by the discriminator. GANs are often used with images and other visual elements and are notoriously efficient in generating compelling and convincing photorealistic images. Most recently, GANs were used to generate an original painting in an unsupervised fashion \citep{artnn}. The following sections go into further detail regarding how the generator and discriminator are trained in GANs. \par

\textit{\textbf{Generator}} - In image synthesis, the generator network can be thought of as a mapping from one representation space (latent space) to another (actual data) \citep{GAN_OV}. When it comes to image synthesis, all of the images in the data space fall into some distribution in a very complex and high-dimensional feature space. Sampling from such a complex space is very difficult, so GANs instead train a generator to create synthetic images from a much more simple feature space (usually random noise) called the latent space. The generator network performs up-sampling of the latent space and is usually a deep neural network consisting of several convolutional and/or fully connected layers \citep{GAN_OV}. The generator is trained using gradient descent to update the weights of the generator network with the aim of producing data (in our case, images) that the discriminator classifies as real. \par
\textit{\textbf{Discriminator}} - The discriminator network can be thought of as a mapping from image data to the probability of the image coming from the real data space, and is also generally a deep neural network consisting of several convolution and/or fully connected layers. However, the discriminator performs down-sampling as opposed to up-sampling. Like the generator, it is trained using gradient descent but its goal is to update the weights so that it is more likely to correctly classify images as real or fake. \par
In GANs, the ideal outcome is for both the generator's and discriminator's cost functions to converge so that the generator produces photo-realistic images that are indistinguishable from real data, and the discriminator at the same time becomes an expert at differentiating between real and synthetic data. This, however, is not possible since a reduction in cost of one model generally leads to an increase in cost of the other. This phenomenon makes training GANs very difficult, and training them simultaneously (both models performing gradient descent in parallel) often leads to a stable orbit where neither model is able to converge. To combat this, the generator and discriminator are often trained independently. In this case, the GAN remains the same, but there are different training stages. In one stage, the weights of the generator are kept constant and gradient descent updates the weights of the discriminator, and in the other stage the weights of the discriminator are kept constant while gradient descent updates the weights of the generator. This is repeated for some number of epochs until a desired low cost for each model is reached \citep{is}. \par

\subsection{cGAN: Conditional GAN}
Conditional Generative Adversarial Networks (cGAN) are an enhancement of GANs proposed by \citet{cgan} shortly after the introduction of GANs by \citet{gan}. The objective function of the cGAN is defined in Eq.~(\ref{eq.cgan.obj}) which is very similar to the GAN objective function in Eq.~(\ref{eq.gan.obj}) except that the inputs to both discriminator and generator are conditioned by a class label $y$.
\begin{equation}
\min_{\theta_g} \max_{\theta_d} V(D_{\theta_d}, G_{\theta_g}) = \mathbb{E}_{x \sim Pdata(x)}[\log(D_{\theta_d}(x|y))] +
\mathbb{E}_{x \sim P_z(z)}[\log(1-D_{\theta_d}(G_{\theta_g}(z|y)))]
\label{eq.cgan.obj}
\end{equation}

The main technical innovation of cGAN is that it introduces an additional input or inputs to the original GAN model, allowing the model to be trained on information such as class labels or other conditioning variables as well as the samples themselves, concurrently. Whereas the original GAN was trained only with samples from the data distribution, resulting in the generated sample reflecting the general data distribution, cGAN enables directing the model to generate more tailored outputs.

\begin{figure}
	\includegraphics[width=\columnwidth]{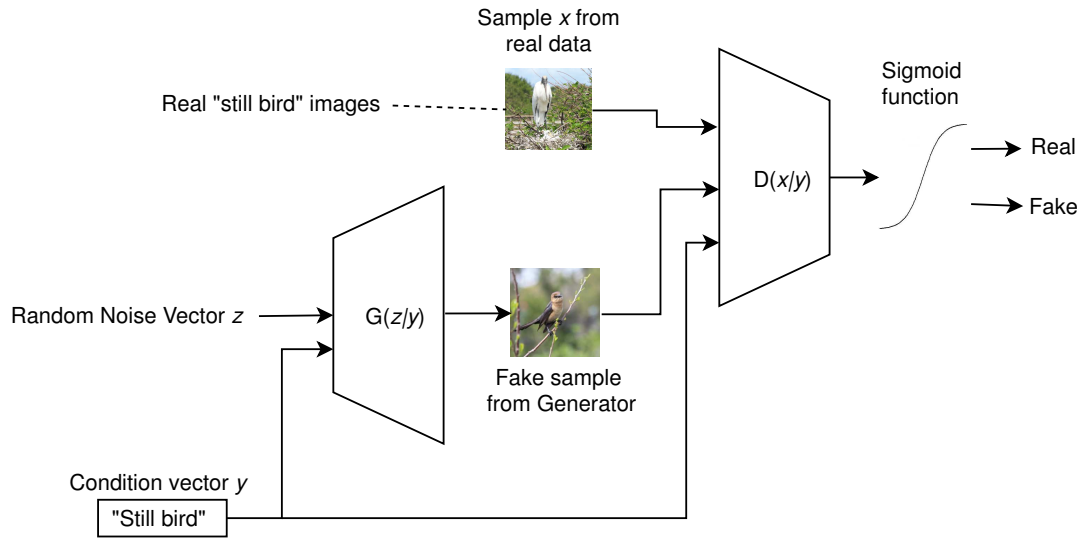}
	\caption{A conceptual view of the conditional GAN architecture. The Generator G(z\(|\)y) generates samples from a random noise distribution and some condition vector (in this case text). The fake samples are fed to the Discriminator D(x\(|\)y) along with real samples and the same condition vector, and the Discriminator calculates the probability that the fake sample came from the real data distribution.} \label{fig.CGAN.structure}
\end{figure}

In Figure \ref{fig.CGAN.structure}, the condition vector is the class label (text string) "Red bird", which is fed to both the generator and discriminator. It is important, however, that the condition vector is related to the real data. If the model in Figure \ref{fig.CGAN.structure} was trained with the same set of real data (red birds) but the condition text was "Yellow fish", the generator would learn to create images of red birds when conditioned with the text "Yellow fish". \par

Note that the condition vector in cGAN can come in many forms, such as texts, not just limited to the class label. Such a unique design provides a direct solution to generate images conditioned by predefined specifications. As a result, cGAN has been used in text-to-image synthesis since the very first day of its invention although modern approaches can deliver much better text-to-image synthesis results.

\begin{figure*}
	\centering
	\includegraphics[width=0.8\columnwidth]{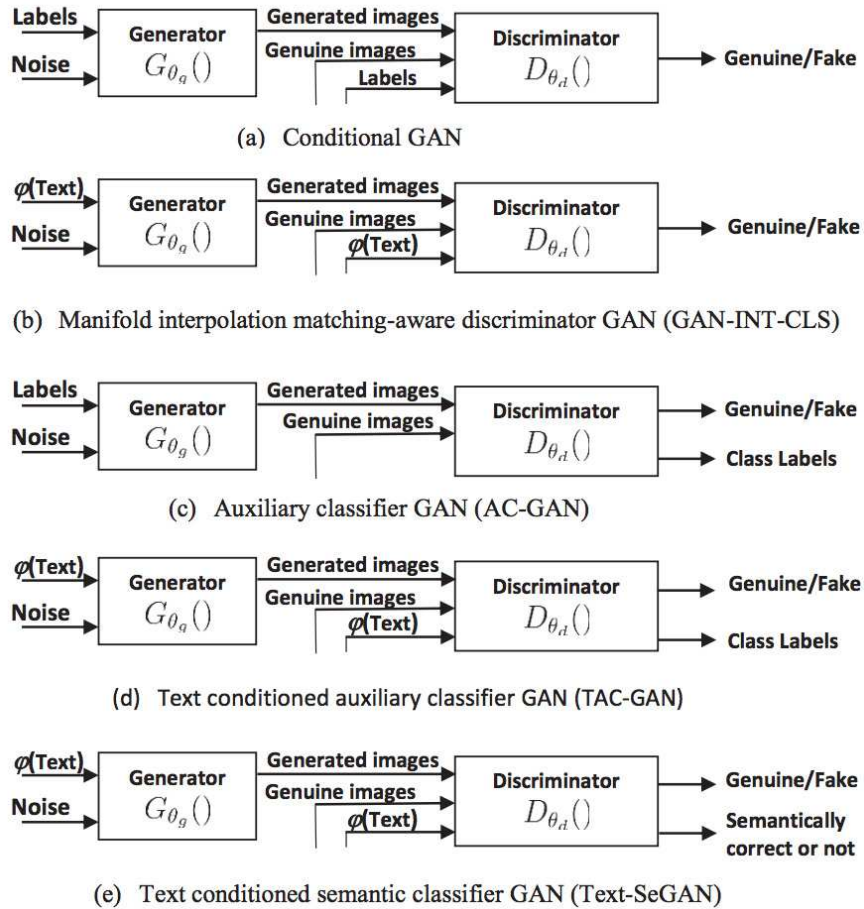}
	\caption{\textcolor{black}{A simple architecture comparisons between five GAN networks for text-to-image synthesis. This figure also explains how texts are fed as input to train GAN to generate images. (a) Conditional GAN (cGAN) \citep{cgan} use labels to condition the input to the generator and the discriminator. The final output is discriminator similar to generic GAN; (b) Manifold interpolation matching-aware discriminator GAN (GAN-INT-CLS) \citep{reed1} feeds text input to both generator and discriminator (texts are preprocessed as embedding features, using function $\varphi()$, and concatenated with other input, before feeding to both generator and discriminator). The final output is discriminator similar to generic GAN; (c) Auxiliary classifier GAN (AC-GAN)~\citep{odena2017conditional} uses an auxiliary classifier layer to predict the class of the image to ensure that the output consists of images from different classes, resulting in diversified synthesis images; (d) text conditioned auxiliary classifier GAN (TAC-GAN) \citep{tacgan} share similar design as GAN-INT-CLS, whereas the output include both a discriminator and a classifier (which can be used for classification); and (e) text conditioned semantic classifier GAN (Text-SeGAN)~\citep{Cha2019AdversarialLO} uses a regression layer to estimate the semantic relevance between the image, so the generated images are not limited to certain classes and are semantically matching to the text input.}}
	\label{fig.gan2image}
\end{figure*}

\textcolor{black}{
	\subsection{Simple GAN Frameworks for Text-to-Image Synthesis}
	In order to generate images from text, one simple solution is to employ the conditional GAN (cGAN) designs and add conditions to the training samples, such that the GAN is trained with respect to the underlying conditions. Several pioneer works have followed similar designs for text-to-image synthesis. }

\textcolor{black}{
	An essential disadvantage of using cGAN for text-to-image synthesis is that that it cannot handle complicated textual descriptions for image generation, because cGAN uses labels as conditions to restrict the GAN inputs. If the text inputs have multiple keywords (or long text descriptions) they cannot be used simultaneously to restrict the input. Instead of using text as conditions, another two approaches~\citep{reed1,tacgan} use text as input features, and concatenate such features with other features to train discriminator and generator, as shown in Figure~\ref{fig.gan2image}(b) and (c). To ensure text being used as GAN input, a feature embedding or feature representation learning \citep{Bengio:pami:2013,zhang:tbig:2018} function $\varphi()$ is often introduced to convert input text as numeric features, which are further concatenated with other features to train GANs.}

\begin{figure*}
	\includegraphics[width=\columnwidth]{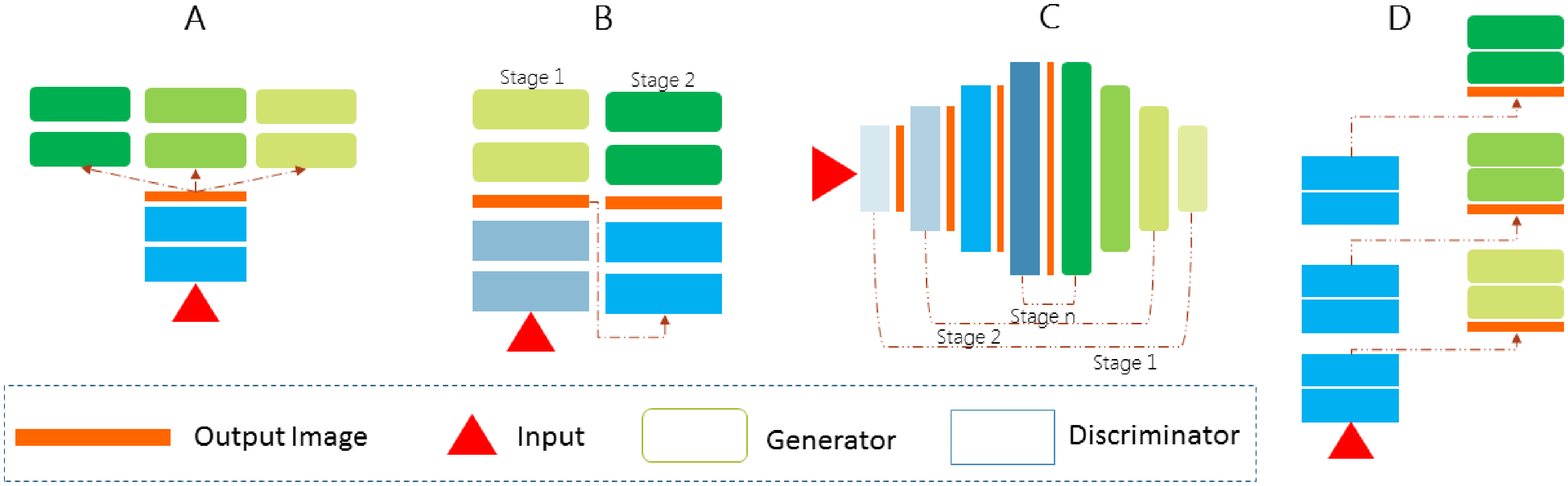}
	\caption{A high level comparison of several advanced GANs framework for text-to-image synthesis. All frameworks take text (red triangle) as input and generate output images. From left to right, (A) uses multiple discriminators and one generator \citep{Durugkar:iclr:2017,nguyen:nips:2017}, (B) uses multiple stage GANs where the output from one GAN is fed to the next GAN as input \citep{stack,NIPS2015_5773}, (C) progressively trains symmetric discriminators and generators \citep{sgan}, and (D) uses a single-stream generator with a hierarchically-nested discriminator trained from end-to-end~\citep{zhang1}. }
	\label{fig.multiple.gan.frameworks}
\end{figure*}

\textcolor{black}{
	\subsection{Advanced GAN Frameworks for Text-to-Image Synthesis}
	Motivated by the GAN and conditional GAN (cGAN) design, many GAN based frameworks have been proposed to generate images, with different designs and architectures, such as using multiple discriminators, using progressively trained discriminators, or using hierarchical discriminators. Figure~\ref{fig.multiple.gan.frameworks} outlines several advanced GAN frameworks in the literature. In addition to these frameworks, many news designs are being proposed to advance the field with rather sophisticated designs. For example, a recent work \citep{gao:AAAI:2019} proposes to use a pyramid generator and three independent discriminators, \textcolor{black}{each focusing on a different aspect of the images, to lead the generator towards creating images that are photo-realistic on multiple levels.} Another recent publication \citep{cha:AAAI:2019} proposes to use discriminator to measure semantic relevance between image and text instead of class prediction (like most discriminator in GANs does), resulting a new GAN structure outperforming text conditioned auxiliary classifier (TAC-GAN) \citep{tacgan} and  generating diverse, realistic, and relevant to the input text regardless of class.}

\textcolor{black}{
	In the following section, we will first propose a taxonomy that summarizes advanced GAN frameworks for text-to-image synthesis, and review most recent proposed solutions to the challenge of generating photo-realistic images conditioned on natural language text descriptions using GANs. The solutions we discuss are selected based on relevance and quality of contributions. Many publications exist on the subject of image-generation using GANs, but in this paper we focus specifically on models for text-to-image synthesis, with the review emphasizing on the ``model'' and ``contributions'' for text-to-image synthesis. At the end of this section, we also briefly review methods using GANs for other image-synthesis applications.}

\begin{figure}
	\centering
	\tiny
	\includegraphics[width=0.8\textwidth]{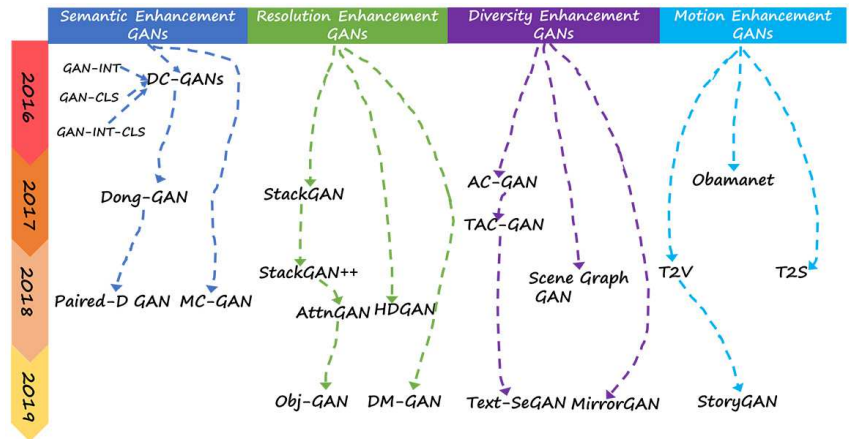}
	\caption{A Taxonomy and categorization of advanced GAN frameworks for Text-to-Image Synthesis. We categorize advanced GAN frameworks into four major categories: Semantic Enhancement GANs, Resolution Enhancement GANs, Diversity Enhancement GANs, and Motion Enhancement GAGs. The relationship between relevant frameworks and their publication date are also outlined as a reference.}
	\label{fig:catGan}
\end{figure}

\textcolor{black}{
	\section{Text-to-Image Synthesis Taxonomy and Categorization}
	In this section, we propose a taxonomy to summarize advanced GAN based text-to-image synthesis frameworks, as shown in Figure~\ref{fig:catGan}. The taxonomy organizes GAN frameworks into four categories, including Semantic Enhancement GANs, Resolution Enhancement GANs, Diversity Enhancement GANs, and Motion Enhancement GAGs. Following the proposed taxonomy, each subsection will introduce several typical frameworks and address their techniques of using GANS to solve certain aspects of the text-to-mage synthesis challenges.}

\textcolor{black}{
	\subsection{GAN based Text-to-Image Synthesis Taxonomy}
	Although the ultimate goal of Text-to-Image synthesis is to generate images closely related to the textual descriptions, the relevance of the images to the texts are often validated from different perspectives, due to the inherent diversity of human perceptions. For example, when generating images matching to the description ``rose flowers'', some users many know the exact type of flowers they like and intend to generate rose flowers with similar colors. Other users, may seek to generate high quality rose flowers with a nice background (\textit{e.g.} garden). The third group of users may be more interested in generating flowers similar to rose but with different colors and visual appearance,  \textit{e.g.}  roses, begonia, and peony. The fourth group of users may want to not only generate flower images, but also use them to form a meaningful action, \textit{e.g.} a video clip showing flower growth, performing a magic show using those flowers, or telling a love story using the flowers.
}


\textcolor{black}{From the text-to-Image synthesis point of view, the first group of users intend to precisely control the semantic of the generated images, and their goal is to match the texts and images at the semantic level. The second group of users are more focused on the resolutions and the qualify of the images, in addition to the requirement that the images and texts are semantically related. For the third group of users, their goal is to diversify the output images, such that their images carry diversified visual appearances and are also semantically related. The fourth user group adds a new dimension in image synthesis, and aims to generate sequences of images which are coherent in temporal order, \textit{i.e.} capture the motion information.}

\textcolor{black}{
	Based on the above descriptions, we categorize GAN based Text-to-Image Synthesis into a taxonomy with four major categories, as shown in Fig. ~\ref{fig:catGan}.
	\begin{itemize}
		\item \textbf{Semantic Enhancement GANs:} Semantic enhancement GANs represent pioneer works of GAN frameworks for text-to-image synthesis. The main focus of the GAN frameworks is to ensure that the generated images are semantically related to the input texts. This objective is mainly achieved by using a neural network to encode texts as dense features, which are further fed to a second network to generate images matching to the texts.
		\item \textbf{Resolution Enhancement GANs:} Resolution enhancement GANs mainly focus on generating high qualify images which are semantically matched to the texts. This is mainly achieved through a multi-stage GAN framework, where the outputs from earlier stage GANs are fed to the second (or later) stage GAN to generate better qualify images.
		\item \textbf{Diversity Enhancement GANs:} Diversity enhancement GANs intend to diversify the output images, such that the generated images are not only semantically related but also have different types and visual appearance. This objective is mainly achieved through an additional component to estimate semantic relevance between generated images and texts, in order to maximize the output diversity.
		\item \textbf{Motion Enhancement GANs:} Motion enhancement GANs intend to add a temporal dimension to the output images, such that they can form meaningful actions with respect to the text descriptions. This goal mainly achieved though a two-step process which first generates images matching to the ``actions'' of the texts, followed by a mapping or alignment procedure to ensure that images are coherent in the temporal order.
	\end{itemize}
}

\textcolor{black}{
	In the following, we will introduce how these GAN frameworks evolve for text-to-image synthesis, and will also review some typical methods of each category.
}

\textcolor{black}{
	\subsection{ Semantic Enhancement GANs}
	Semantic relevance is one the of most important criteria of the text-to-image synthesis. For most GNAs discussed in this survey, they are required to generate images semantically related to the text descriptions. However, the semantic relevance is a rather subjective measure, and images are inherently rich in terms of its semantics and interpretations. Therefore, many GANs are further proposed to enhance the text-to-image synthesis from different perspectives. In this subsection, we will review several classical approaches which are commonly served as text-to-image synthesis baseline.
}

\textcolor{black}{
	\subsubsection{DC-GAN}
	Deep convolution generative adversarial network (DC-GAN)~\citep{reed1} represents the pioneer work for text-to-image synthesis using GANs. Its main goal is to train a deep convolutional generative adversarial network (DC-GAN) on text features. During this process these text features are encoded by another neural network. This neural network is a hybrid convolutional recurrent network at the character level. Concurrently, both neural networks have also feed-forward inference in the way they condition text features. Generating realistic images automatically from natural language text is the motivation of several of the works proposed in this computer vision field. However, actual artificial intelligence (AI) systems are far from achieving this task~\citep{reed1,cycgen,multitask,txt2txt,s2,sur3,cgan}. Lately, recurrent neural networks led the way to develop frameworks that learn discriminatively on text features. At the same time, generative adversarial networks (GANs) began recently to show some promise on generating compelling images of a whole host of elements including but not limited to faces, birds, flowers, and non-common images such as room interiors\citep{reed1}. DC-GAN is a multimodal learning model that attempts to bridge together both of the above mentioned unsupervised machine learning algorithms, the recurrent neural networks (RNN) and generative adversarial networks (GANs), with the sole purpose of speeding the generation of text-to-image synthesis.}

\textcolor{black}{
	Deep learning shed some light to some of the most sophisticated advances in natural language representation, image synthesis \citep{sur1,reed1,ganinout,sgan}, and classification of generic data \citep{han:IS:2019}. However, a bulk of the latest breakthroughs in deep learning and computer vision were related to supervised learning~\citep{reed1}. Even though natural language and image synthesis were part of several contributions on the supervised side of deep learning, unsupervised learning saw recently a tremendous rise in input from the research community specially on two subproblems: text-based natural language and image synthesis~\citep{i2t2i,lrgan,reed1,cha1,stackv2}. These subproblems are typically subdivided as focused research areas. DC-GAN's contributions are mainly driven by these two research areas. In order to generate plausible images from natural language, DC-GAN contributions revolve around developing a straightforward yet effective GAN architecture and training strategy that allows natural text to image synthesis. These contributions are primarily tested on the Caltech-UCSD Birds and Oxford-102 Flowers datasets. Each image in these datasets carry five text descriptions. These text descriptions were created by the research team when setting up the evaluation environment. The DC-GANs model is subsequently trained on several subcategories. Subcategories in this research represent the training and testing sub datasets. The performance shown by these experiments display a promising yet effective way to generate images from textual natural language descriptions~\citep{reed1}.}

\textcolor{black}{
	\subsubsection{DC-GAN Extensions}
	Following the pioneer DC-GAN framework~\citep{reed1}, many researches propose revised network structures (\textit{e.g.} different discriminaotrs) in order to improve images with better semantic relevance to the texts. Based on the deep convolutional adversarial network (DC-GAN) network architecture, GAN-CLS with image-text matching discriminator, GAN-INT learned with text manifold interpolation and GAN-INT-CLS which combines both are proposed to find semantic match between text and image. Similar to the DC-GAN architecture,  an adaptive loss function (\textit{i.e.} Perceptual Loss~\citep{johnson2016perceptual}) is proposed for semantic image synthesis which can synthesize a realistic image that not only matches the target text description but also keep the irrelavant features(e.g. background) from source images~\citep{dong2017semantic}. Regarding to the Perceptual Losses, three loss functions (\textit{i.e.} Pixel reconstruction loss, Activation reconstruction loss and Texture reconstruction loss) are proposed in~\citep{cha2017adversarial} in which they construct the network architectures based on the DC-GAN, \textit{i.e.} GAN-INT-CLS-Pixel, GAN-INT-CLS-VGG and GAN-INT-CLS-Gram with respect to three losses. In \citep{dong2017semantic}, a residual transformation unit is added in the network to retain similar structure of the source image.}

\textcolor{black}{
	Following the \citep{dong2017semantic} and considering the features in early layers address background while
	foreground is obtained in latter layers in CNN, a pair of discriminators with different architectures (\textit{i.e.} Paired-D GAN) is proposed to synthesize background and foreground from a source image seperately~\citep{vo2018paired}. Meanwhile, the skip-connection in the generator is employed to more precisely retain background information in the source image.}

\textcolor{black}{
	\subsubsection{MC-GAN}
	When synthesising images, most text-to-image synthesis methods consider each output image as one single unit to characterize its semantic relevance to the texts. This is likely problematic because most images naturally consist of two crucial components: foreground and background. Without properly separating these two components, it's hard to characterize the semantics of an image if the whole image is treated as a single unit without proper separation.}

\textcolor{black}{
	In order to enhance the semantic relevance of the images, a multi-conditional GAN (MC-GAN)~\citep{park2018mc} is proposed to  synthesize a target image by combining the background of a source image and a text-described foreground object which does not exist in the source image. A unique feature of MC-GAN is that it proposes a synthesis block in which the background  feature is extracted from the given image without non-linear function (\textit{i.e.} only using convolution and batch normalization) and the foreground  feature is the feature map from the previous layer.
}

\textcolor{black}{
	Because MC-GAN is able to properly model the background and foreground of the generated images, a unique strength of MC-GAN is that users are able to provide a base image and MC-GAN is able to preserve the background information of the base image to generate new images.
}
\textcolor{black}{
	\subsection{Resolution Enhancement GANs}
	Due to the fact that training GANs will be much difficult when generating high-resolution images, a two stage GAN (\textit{i.e.} stackGAN) is proposed in which rough images(\textit{i.e.} low-resolution images) are generated in stage-I and refined in stage-II. To further improve the quality of generated images, the second version of StackGAN (\textit{i.e.} Stack++) is proposed to use multi-stage GANs to generate multi-scale images.   A color-consistency regularization term is also added into the loss to keep the consistency of images in different scales. }

\textcolor{black}{
	While stackGAN and StackGAN++ are both built on the global sentence vector, AttnGAN is proposed to use attention mechanism (\textit{i.e.} Deep Attentional Multimodal Similarity Model (DAMSM)) to model the multi-level information (\textit{i.e.} word level and sentence level) into GANs. In the following, StackGAN, StackGAN++ and AttnGAN will be explained in detail. }

\textcolor{black}{
	Recently, Dynamic Memory Generative Adversarial Network (\textit{i.e.} DM-GAN)\citep{zhu2019dm}  which uses a dynamic memory component is proposed to focus on refiningthe initial generated image which is the key to the success of generating high quality images.
}
\subsubsection{StackGAN}
In 2017, Zhang et al. proposed a model for generating photo-realistic images from text descriptions called StackGAN (Stacked Generative Adversarial Network) \citep{stack}. In their work, they define a two-stage model that uses two cascaded GANs, each corresponding to one of the stages. The stage I GAN takes a text description as input, converts the text description to a text embedding containing several conditioning variables, and generates a low-quality 64x64 image with rough shapes and colors based on the computed conditioning variables. The stage II GAN then takes this low-quality stage I image as well as the same text embedding and uses the conditioning variables to correct and add more detail to the stage I result. The output of stage II is a photorealistic 256$times$256 image that resembles the text description with compelling accuracy. \par

One major contribution of StackGAN is the use of cascaded GANs for text-to-image synthesis through a sketch-refinement process. By conditioning the stage II GAN on the image produced by the stage I GAN and text description, the stage II GAN is able to correct defects in the stage I output, resulting in high-quality 256x256 images. Prior works have utilized ``stacked'' GANs to separate the image generation process into structure and style \citep{s2}, multiple stages each generating lower-level representations from higher-level representations of the previous stage \citep{sgan}, and multiple stages combined with a laplacian pyramid approach \citep{lpyr2}, which was introduced for image compression by P. Burt and E. Adelson in 1983 and uses the differences between consecutive down-samples of an original image to reconstruct the original image from its down-sampled version \citep{lpyr1}. However, these works did not use text descriptions to condition their generator models. \par

Conditioning Augmentation is the other major contribution of StackGAN. Prior works transformed the natural language text description into a fixed text embedding containing static conditioning variables which were fed to the generator \citep{reed1}. StackGAN does this and then creates a Gaussian distribution from the text embedding and randomly selects variables from the Gaussian distribution to add to the set of conditioning variables during training. This encourages robustness by introducing small variations to the original text embedding for a particular training image while keeping the training image that the generated output is compared to the same. The result is that the trained model produces more diverse images in the same distribution when using Conditioning Augmentation than the same model using a fixed text embedding \citep{stack}.

\subsubsection{StackGAN++}
Proposed by the same users as StackGAN, StackGAN++ is also a stacked GAN model, but organizes the generators and discriminators in a ``tree-like'' structure \citep{stackv2} with multiple stages. The first stage combines a noise vector and conditioning variables (with Conditional Augmentation introduced in \citep{stack}) for input to the first generator, which generates a low-resolution image, 64$\times$64 by default (this can be changed depending on the desired number of stages). Each following stage uses the result from the previous stage and the conditioning variables to produce gradually higher-resolution images. These stages do not use the noise vector again, as the creators assume that the randomness it introduces is already preserved in the output of the first stage. The final stage produces a 256$\times$256 high-quality image. \par

StackGAN++ introduces the joint conditional and unconditional approximation in their designs \citep{stackv2}. The discriminators are trained to calculate the loss between the image produced by the generator and the conditioning variables (measuring how accurately the image represents the description) as well as the loss between the image and real images (probability of the image being real or fake). The generators then aim to minimize the sum of these losses, improving the final result. \par

\subsubsection{AttnGAN}
Attentional Generative Adversarial Network (AttnGAN)~\citep{attn} is very similar, in terms of its structure, to StackGAN++ \citep{stackv2}, discussed in the previous section, but some novel components are added. Like previous works \citep{reed2, reed1, stack, stackv2}, a text encoder generates a text embedding with conditioning variables based on the overall sentence. Additionally, the text encoder generates a separate text embedding with conditioning variables based on individual words. This process is optimized to produce meaningful variables using a bidirectional recurrent neural network (BRNN), more specifically bidirectional Long Short Term Memory (LSTM) \citep{bdrnn}, which, for each word in the description, generates conditions based on the previous word as well as the next word (bidirectional). The first stage of AttnGAN generates a low-resolution image based on the sentence-level text embedding and random noise vector. The output is fed along with the word-level text embedding to an ``attention model'', which matches the word-level conditioning variables to regions of the stage I image, producing a word-context matrix. This is then fed to the next stage of the model along with the raw previous stage output. Each consecutive stage works in the same manner, but produces gradually higher-resolution images conditioned on the previous stage. \par

Two major contributions were introduced in AttnGAN: the attentional generative network and the Deep Attentional Multimodal Similarity Model (DAMSM) \citep{stackv2}. The attentional generative network matches specific regions of each stage's output image to conditioning variables from the word-level text embedding. This is a very worthy contribution, allowing each consecutive stage to focus on specific regions of the image independently, adding ``attentional'' details region by region as opposed to the whole image. The DAMSM is also a key feature introduced by AttnGAN, which is used after the result of the final stage to calculate the similarity between the generated image and the text embedding at both the sentence level and the more fine-grained word level. Table \ref{tab2} shows scores from different metrics for StackGAN, StackGAN++, AttnGAN, and HDGAN on the CUB, Oxford, and COCO datasets. The table shows that AttnGAN outperforms the other models in terms of IS on the CUB dataset by a small amount and greatly outperforms them on the COCO dataset.

\subsubsection{HDGAN}
Hierarchically-nested adversarial network (HDGAN) is a method proposed by \citet{zhang1}, and its main objective is to tackle the difficult problem of dealing with photographic images from semantic text descriptions. These semantic text descriptions are applied on images from diverse datasets. This method introduces adversarial objectives nested inside hierarchically oriented networks~\citep{zhang1}. Hierarchical networks helps regularize mid-level manifestations. In addition to regularize mid-level manifestations, it assists the training of the generator in order to capture highly complex still media elements. These elements are captured in statistical order to train the generator based on settings extracted directly from the image. The latter is an ideal scenario. However, this paper aims to incorporate a single-stream architecture. This single-stream architecture functions as the generator that will form an optimum adaptability towards the jointed discriminators. Once jointed discriminators are setup in an optimum manner, the single-stream architecture will then advance generated images to achieve a much higher resolution~\citep{zhang1}.

The main contributions of the HDGANs include the introduction of a visual-semantic similarity measure~\citep{zhang1}. This feature will aid in the evaluation of the consistency of generated images. In addition to checking the consistency of generated images, one of the key objectives of this step is to test the logical consistency of the end product~\citep{zhang1}. The end product in this case would be images that are semantically mapped from text-based natural language descriptions to each area on the picture \textit{e.g.} a wing on a bird or petal on a flower. Deep learning has created a multitude of opportunities and challenges for researchers in the computer vision AI field. Coupled with GAN and multimodal learning architectures, this field has seen tremendous growth~\citep{reed1,cycgen,multitask,txt2txt,s2,sur3,cgan}. Based on these advancements, HDGANs attempt to further extend some desirable and less common features when generating images from textual natural language~\citep{zhang1}. In other words, it takes sentences and treats them as a hierarchical structure. This has some positive and negative implications in most cases. For starters, it makes it more complex to generate compelling images. However, one of the key benefits of this elaborate process is the realism obtained once all processes are completed. In addition, one common feature added to this process is the ability to identify parts of sentences with bounding boxes. If a sentence includes common characteristics of a bird, it will surround the attributes of such bird with bounding boxes. In practice, this should happen if the desired image have other elements such as human faces (\textit{e.g.} eyes, hair, etc), flowers (\textit{e.g.} petal size, color, etc), or any other inanimate object (\textit{e.g.} a table, a mug, etc). Finally, HDGANs evaluated some of its claims on common ideal text-to-image datasets such as CUB, COCO, and Oxford-102~\citep{reed1,zhang1,cycgen,multitask,txt2txt,s2,sur3,cgan}. These datasets were first utilized on earlier works \citep{reed1}, and most of them sport modified features such image annotations, labels, or descriptions. The qualitative and quantitative results reported by researchers in this study were far superior of earlier works in this same field of computer vision AI.


\textcolor{black}{
	\subsection{Diversity Enhancement GANs}
	In this subsection, we introduce text-to-image synthesis methods which try to maximize the diversity of the output images, based on the text descriptions.
}

\textcolor{black}
{
	\subsubsection{AC-GAN}
	Two issues arise in the traditional GANs~\citep{mirza2014conditional} for image synthesis: (1) scalabilirty problem: traditional GANs cannot predict a large number of image categories; and (2) diversity problem: images are often subject to one-to-many mapping, so one image could be labeled as different tags or being described using different texts. To address these problems, GAN conditioned on additional information, \textit{e.g.} cGAN, is an alternative solution. However, although cGAN and many previously introduced approaches are able to generate images with respect to the text descriptions, they often output images with similar types and visual appearance.
}

\textcolor{black}
{
	Slightly different from the cGAN, auxiliary classifier GANs (AC-GAN) \citep{odena2017conditional} proposes to improve the diversity of output images by using an auxiliary classifier to control output images. The overall structure of AC-GAN is shown in Fig.~\ref{fig.gan2image}(c). In AC-GAN, every generated image is associated with a class label, in addition to the true/fake label which are commonly used in GAN or cGAN. The discriminator of AC-GAN not only outputs a probability distribution over sources (\textit{i.e.} whether the image is true or fake), it also output a probability distribution over the class label (\textit{i.e.} predict which class the image belong to).}

\textcolor{black}
{
	By using an auxiliary classifier layer to predict the class of the image, AC-GAN is able to use the predicted class labels of the images to ensure that the output consists of images from different classes, resulting in diversified synthesis images. The results show that AC-GAN can generate images with high diversity.
}

\textcolor{black}{
	\subsubsection{TAC-GAN}
	Building on the AC-GAN, TAC-GAN~\citep{dash2017tac} is proposed to replace the class information with textual descriptions as the input to perform the task of text to image synthesis. The architecture of TAC-GAN is shown in Fig.~\ref{fig.gan2image}(d), which is similar to AC-GAN. Overall, the major difference between TAC-GAN and AC-GAN is that TAC-GAN conditions the generated images on text descriptions instead of on a class label. This design makes TAC-GAN more generic for image synthesis.
}

\textcolor{black}{
	For TAC-GAN, it imposes restrictions on generated images in both texts and class labels. The input vector of TAC-GAN's
	generative network is built based on a noise vector and embedded vector representation of textual descriptions. The discriminator of TAC-GAN is similar to that of the AC-GAN, which not only predicts whether the image is fake or not, but also predicts the label of the images. A minor difference of TAC-GAN's discriminator, compared to that of the AC-GAN, is that it also receives text
	information as input before performing its classification.
}

\textcolor{black}
{
	The experiments and validations, on the Oxford-102 flowers dataset, show that the results produced by TAC-GAN are ``slightly better'' that other approaches, including GAN-INT-CLS and StackGAN.
}

\textcolor{black}{
	\subsubsection{Text-SeGAN}
	In order to improve the diversity of the output images, both AC-GAN and TAC-GAN's discriminators predict class labels of the synthesised images. This process likely enforces the semantic diversity of the images, but class labels are inherently restrictive in describing image semantics, and images described by text can be matched to multiple labels. Therefore, instead of predicting images' class labels, an alternative solution is to directly quantify their semantic relevance.
}

\textcolor{black}
{
	The architecture of Text-SeGAN is shown in Fig.~\ref{fig.gan2image}(e). In order to directly quantify semantic relevance, Text-SeGAN~\citep{Cha2019AdversarialLO} adds a regression layer to estimate the semantic relevance between the image and text instead of a classifier layer of predicting labels. The estimated semantic reference is a fractional value ranging between 0 and 1, with a higher value reflecting better semantic relevance between the image and text. Due to this unique design, an inherent advantage of Text-SeGAN is that the generated images are not limited to certain classes and are semantically matching to the text input.
}

\textcolor{black}{
	Experiments and validations, on Oxford-102 flower dataset, show that Text-SeGAN can generate diverse images that are semantically relevant to the input text. In addition, the results of Text-SeGAN show improved inception score compared to other approaches, including GAN-INT-CLS, StackGAN, TAC-GAN, and HDGAN.
}

\textcolor{black}
{
	\subsubsection{MirrorGAN and Scene Graph GAN}
	Due to the inherent complexity of the visual images, and the diversity of text descriptions (\textit{i.e.} same words could imply different meanings), it is difficulty to precisely match the texts to the visual images at the semantic levels. For most methods we have discussed so far, they employ a direct text to image generation process, but there is no validation about how generated images comply with the text in a reverse fashion.
}

\textcolor{black}
{
	To ensure the semantic consistency and diversity, MirrorGAN~\citep{DBLP:journals/corr/abs-1903-05854} employs a mirror structure, which reversely learns from generated images to output texts (an image-to-text process) to further validate whether generated are indeed consistent to the input texts. MirrowGAN includes three modules: a semantic text embedding module (STEM), a global-local collaborative attentive module for cascaded image generation (GLAM), and a semantic text regeneration and alignment module (STREAM). The back to back Text-to-Image (T2I) and Image-to-Text (I2T) are combined to  progressively enhance the diversity and semantic consistency of the generated images.
}

\textcolor{black}
{
	In order to enhance the diversity of the output image, Scene Graph GAN~\citep{johnson2018cvpr} proposes to use visual scene graphs to describe the layout of the objects, allowing users to precisely specific the relationships between objects in the images. In order to convert the visual scene graph as input for GAN to generate images, this method uses graph convolution to process input graphs. It computes a scene layout by predicting bounding boxes and segmentation masks for objects. After that, it converts the computed layout to an image with a cascaded reﬁnement network.
}

\textcolor{black}{
	\subsection{Motion Enhancement GANs}
	Instead of focusing on generating static images, another line of text-to-image synthesis research focuses on generating videos (\textit{i.e.} sequences of images) from texts. In this context, the synthesised videos are often useful resources for automated assistance or story telling.}

\textcolor{black}{
	\subsubsection{ObamaNet and T2S}
	One early/interesting work of motion enhancement GANs is to generate spoofed speech and lip-sync videos (or talking face) of Barack Obama (\textit{i.e.} ObamaNet) based on text input~\citep{kumar2017obamanet}. This framework is consisted of three parts, \textit{i.e.} text to speech using ``Char2Wav'', mouth shape representation synced to the audio using a time-delayed LSTM and ``video generation'' conditioned on the mouth shape using ``U-Net'' architecture. Although the results seem promising, ObamaNet only models the mouth region and the videos are not generated from noise which can be regarded as video prediction other than video generation. }

\textcolor{black}{
	Another meaningful trial of using synthesised videos for automated assistance is to translate spoken language (\textit{e.g.} text) into sign language video sequences (\textit{i.e.} T2S)~\citep{stoll2018sign}. This is often achieved through a two step process: converting texts as meaningful units to generate images, followed by a learning component to arrange images into sequential order for best representation. More specifically, using RNN based machine translation methods, texts are translated into  sign language gloss sequences. Then, glosses are mapped to skeletal pose sequences using a lookup-table. To generate videos, a conditional DCGAN with the input of concatenation of latent representation of the image for a base pose and skeletal pose information is built.
}

\textcolor{black}{
	\subsubsection{T2V}
	In \citet{li2018video}, a text-to-video model (\textit{T2V}) is proposed based on the \textit{cGAN} in which the input is the isometric Gaussian noise with the text-gist vector served as the generator. A key component of generating videos from text is to train a conditional generative model to extract both static
	and dynamic information from text, followed by a hybrid framework combining a Variational Autoencoder (VAE) and a Generative Adversarial Network (GAN).
}

\textcolor{black}{
	More specifically, T2V relies on two types of features, static features and dynamic features, to generate videos. Static features, called ``gist'' are used to sketch text-conditioned background color and object layout structure. Dynamic features, on the other hand, are considered by transforming input text into an image filter which eventually forms the video generator which consists of three entangled neural networks. The text-gist vector is generated by a gist generator which maintains static information (\textit{e.g.} background) and a text2filter which  captures the dynamic information (\textit{i.e.} actions) in the text to generate videos.
}

\textcolor{black}{
	As demonstrated in the paper~\citep{li2018video}, the generated videos are semantically related to the texts, but have a rather low quality (\textit{e.g.} only $64 \times 64$ resolution).
}

\textcolor{black}{
	\subsubsection{StoryGAN}
	Different from \textit{T2V} which generates videos from a single text, \textit{StoryGAN} aims to produce dynamic scenes consistent of specified texts (\textit{i.e.} story written in a multi-sentence paragraph) using a sequential GAN model~\citep{li2019storygan}.  Story encoder, context  encoder, and discriminators are the main components of this model. By using stochastic sampling, the story encoder intends to learn an low-dimensional embedding vector for the whole story to keep the continuity of the story. The context  encoder is proposed to capture contextual information during sequential image generation based on a deep RNN. Two discriminators of StoryGAN are image discriminator which evaluates the generated images and story discriminator which ensures the global consistency. }

\textcolor{black}
{
	The experiments and comparisons, on CLEVR dataset and Pororo cartoon dataset which are originally used for visual question answering, show that StoryGAN improves the generated video qualify in terms of Structural Similarity Index (SSIM), visual qualify, consistence, and relevance (the last three measure are based on human evaluation).
}



\begin{table}[htbp]
	\begin{center}
		\caption{A summary of different GANs and datasets used for validation. A $\checkmark$ symbol indicates that the model was evaluated using the corresponding dataset}
		\begin{tabular}{|c|c|c|c|c|c|}
			\hline
			\textbf{Method}&\multicolumn{5}{|c|}{\textbf{Evaluation Datasets}} \\
			\cline{2-6}
			\textbf{Names} & \textbf{\textit{MNIST}}& \textbf{\textit{Oxford-102}}& \textbf{\textit{COCO}}& \textbf{\textit{CUB}} & \textbf{\textit{CIFAR-10}} \\
			\hline
			cGAN \citep{cgan} & \checkmark &  &  & &  \\
			AC-GAN \citep{odena2017conditional} & &  &  &  & \checkmark \\
			TAC-GAN \citep{tacgan} & & \checkmark &  &  &  \\
			Text-SeGAN \citep{Cha2019AdversarialLO} & & \checkmark &  &  & \\
			GAN-INT-CLS \citep{reed1}& & \checkmark & \checkmark & \checkmark & \\
			StackGAN \citep{stack} & & \checkmark & \checkmark & \checkmark &\\
			StackGAN++ \citep{stackv2} & & \checkmark & \checkmark & \checkmark & \\
			AttnGAN \citep{attn}& & \checkmark & \checkmark & \checkmark &\\
			DC-GAN \citep{reed1} & & \checkmark & \checkmark & \checkmark & \\
			HDGAN \citep{zhang1}& & \checkmark & \checkmark & \checkmark  &\\
			MirrorGAN \citep{DBLP:journals/corr/abs-1903-05854}& &  & \checkmark & \checkmark  &\\
			\hline

		\end{tabular}
		\bigskip
		\label{tab1}
	\end{center}
\end{table}

\section{GAN Based Text-to-Image Synthesis Applications, Benchmark, and Evaluation and Comparisons}

\subsection{Text-to-image Synthesis Applications}
Computer vision applications have strong potential for industries including but not limited to the medical, government, military, entertainment, and online social media fields \citep{sur1,med,hong1,gifcwa,maocv, wla}. Text-to-image synthesis is one such application in computer vision AI that has become the main focus in recent years due to its potential for providing beneficial properties and opportunities for a wide range of applicable areas. \par

Text-to-image synthesis is an application byproduct of deep convolutional decoder networks in combination with GANs \citep{sur1,reed1,attn}. Deep convolutional networks have contributed to several breakthroughs in image, video, speech, and audio processing. This learning method intends, among other possibilities, to help translate sequential text descriptions to images supplemented by one or many additional methods. Algorithms and methods developed in the computer vision field have allowed researchers in recent years to create realistic images from plain sentences. Advances in the computer vision, deep convolutional nets, and semantic units have shined light and redirected focus to this research area of text-to-image synthesis, having as its prime directive: to aid in the generation of compelling images with as much fidelity to text descriptions as possible. \par

To date, models for generating synthetic images from textual natural language in research laboratories at universities and private companies have yielded compelling images of flowers and birds \citep{reed1}. Though flowers and birds are the most common objects studied thus far, research has been applied to other classes as well. For example, there have been studies focused solely on human faces \citep{sur1,reed1,t2s,semilat}. \par

It’s a fascinating time for computer vision AI and deep learning researchers and enthusiasts. The consistent advancement in hardware, software, and contemporaneous development of computer vision AI research disrupts multiple industries. These advances in technology allow for the extraction of several data types from a variety of sources. For example, image data captured from a variety of photo-ready devices, such as smart-phones, and online social media services opened the door to the analysis of large amounts of media datasets \citep{wla}. The availability of large media datasets allow new frameworks and algorithms to be proposed and tested on real-world data. \par

\subsection{Text-to-image Synthesis Benchmark Datasets}
A summary of some reviewed methods and benchmark datasets used for validation is reported in Table~\ref{tab1}. In addition, the performance of different GANs with respect to the benchmark datasets and performance metrics is reported in Table~\ref{tab2}. \par

In order to synthesize images from text descriptions, many frameworks have taken a minimalistic approach by creating small and background-less images \citep{lstsq}. In most cases, the experiments were conducted on simple datasets, initially containing images of birds and flowers. \citet{reed1} contributed to these data sets by adding corresponding natural language text descriptions to subsets of the CUB, MSCOCO, and Oxford-102 datasets, which facilitated the work on text-to-image synthesis for several papers released more recently. \par

While most deep learning algorithms use MNIST \citep{lecun-mnisthandwrittendigit-2010} dataset as the benchmark, there are three main datasets that are commonly used for evaluation of proposed GAN models for text-to-image synthesis: CUB \citep{cub}, Oxford \citep{ox}, COCO \citep{coco}, and CIFAR-10~\citep{cifar-10}. CUB \citep{cub} contains 200 birds with matching text descriptions and Oxford \citep{ox} contains 102 categories of flowers with 40-258 images each and matching text descriptions. These datasets contain individual objects, with the text description corresponding to that object, making them relatively simple. COCO \citep{coco} is much more complex, containing 328k images with 91 different object types. CIFAI-10~\citep{cifar-10} dataset consists of 60000 32$times$32 colour images in 10 classes, with 6000 images per class. In contrast to CUB and Oxford, whose images each contain an individual object, COCO’s images may contain multiple objects, each with a label, so there are many labels per image. The total number of labels over the 328k images is 2.5 million \citep{coco}. \par

\subsection{Text-to-image Synthesis Benchmark Evaluation Metrics}
Several evaluation metrics are used for judging the images produced by text-to-image GANs. Proposed by \citet{is}, Inception Scores (IS) calculates the entropy (randomness) of the conditional distribution, obtained by applying the Inception Model introduced in \citep{im}, and marginal distribution of a large set of generated images, which should be low and high, respectively, for meaningful images. Low entropy of conditional distribution means that the evaluator is confident that the images came from the data distribution, and high entropy of the marginal distribution means that the set of generated images is diverse, which are both desired features. The IS score is then computed as the KL-divergence between the two entropies. FCN-scores \citep{fcn} are computed in a similar manner, relying on the intuition that realistic images generated by a GAN should be able to be classified correctly by a classifier trained on real images of the same distribution. Therefore, if the FCN classifier classifies a set of synthetic images accurately, the image is probably realistic, and the corresponding GAN gets a high FCN score. Frechet Inception Distance (FID) \citep{fid} is the other commonly used evaluation metric, and takes a different approach, actually comparing the generated images to real images in the distribution. A high FID means there is little relationship between statistics of the synthetic and real images and vice versa, so lower FIDs are better.

\begin{figure}[ht]
	\centering
	\includegraphics[width=0.8\textwidth]{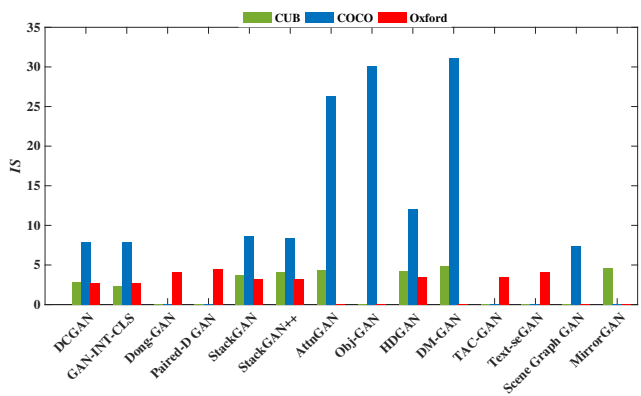}
	\caption{Performance comparison between 14 GANs with respect to their Inception Scores (IS).}
	\label{fig:ISMeasure}
\end{figure}

\textcolor{black}
{
	The performance of different GANs with respect to the benchmark datasets and performance metrics is reported in Table~\ref{tab2}. In addition, Figure~\ref{fig:ISMeasure} further lists the performance of 14 GANs with respect to their Inception Scores (IS).
}

\subsection{GAN Based Text-to-image Synthesis Results Comparison}
\begin{figure}[!ht]
	\centering
	\includegraphics[width=0.8\columnwidth]{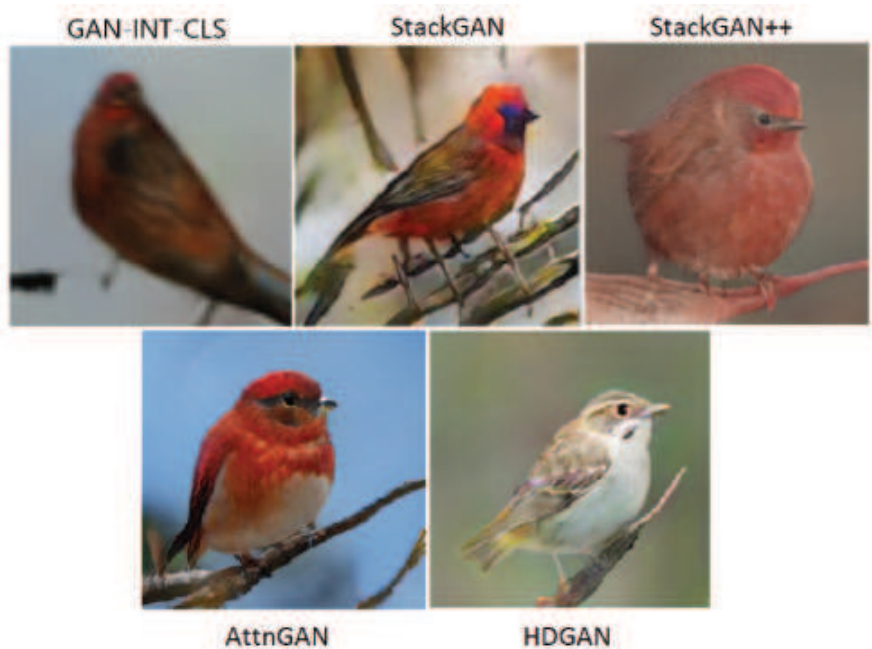}
	\caption{Examples of best images of ``birds'' generated by GAN-INT-CLS, StackGAN, StackGAN++, AttnGAN, and HDGAN. Images reprinted from \citet{stack,stack,stackv2,attn}, and \citet{zhang1}, respectively.} 
	\label{birds}
\end{figure}

\begin{figure}[!ht]
	\centering
	\includegraphics[width=0.8\columnwidth]{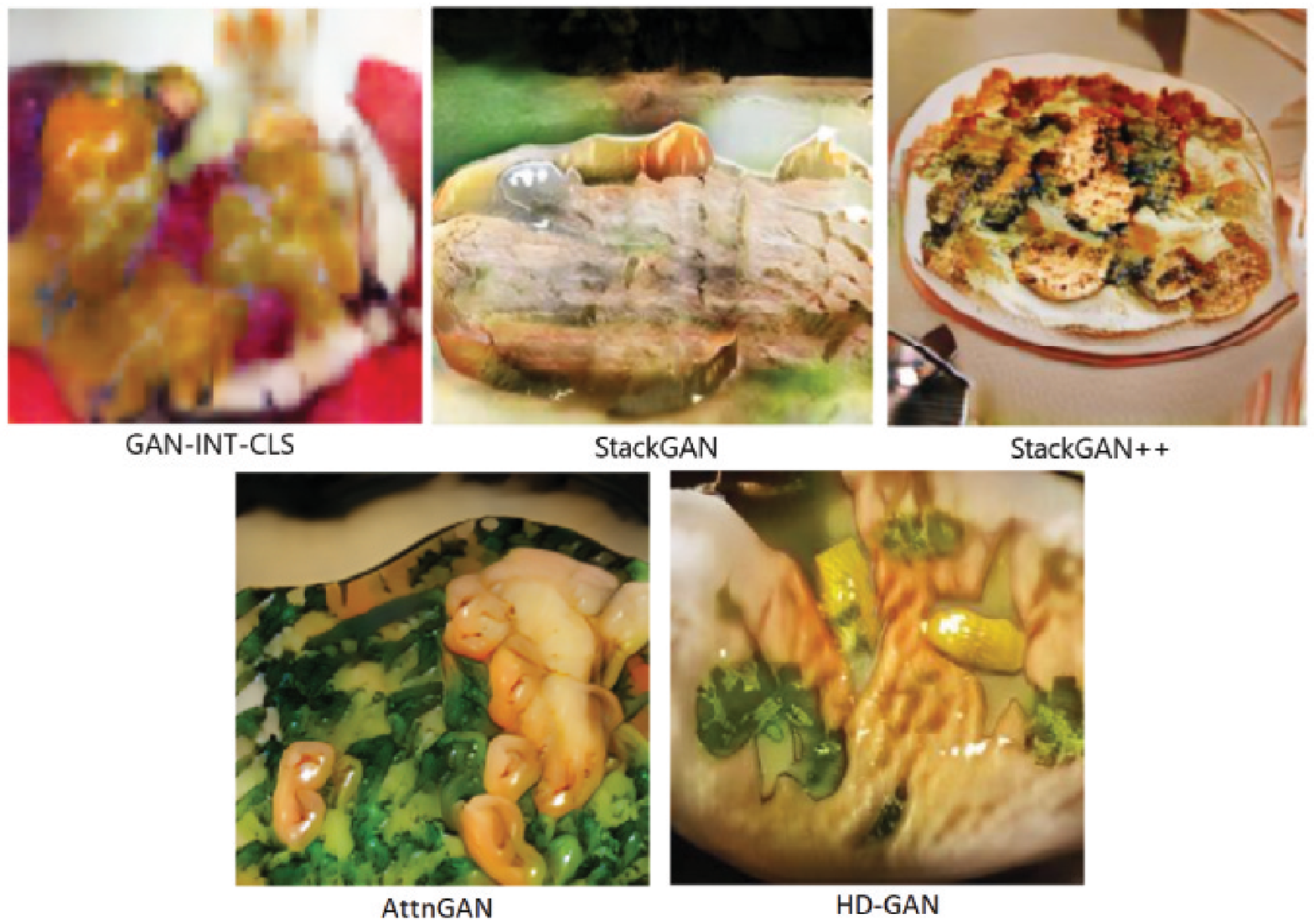}
	\caption{Examples of best images of ``a plate of vegetables'' generated by GAN-INT-CLS, StackGAN, StackGAN++, AttnGAN, and HDGAN. Images reprinted from \citet{stack,stack,stackv2,attn}, and \citet{zhang1}, respectively.} 
	\label{vegetables}
\end{figure}

While we gathered all the data we could find on scores for each model on the CUB, Oxford, and COCO datasets using IS, FID, FCN, and human classifiers, we unfortunately were unable to find certain data for AttnGAN and HDGAN (missing in Table \ref{tab2}). The best evaluation we can give for those with missing data is our own opinions by looking at examples of generated images provided in their papers. In this regard, we observed that HDGAN produced relatively better visual results on the CUB and Oxford datasets while AttnGAN produced far more impressive results than the rest on the more complex COCO dataset. This is evidence that the attentional model and DAMSM introduced by AttnGAN are very effective in producing high-quality images. Examples of the best results of birds and plates of vegetables generated by each model are presented in Figures \ref{birds} and \ref{vegetables}, respectively. \par


\textcolor{black}{In terms of inception score (IS), which is the metric that was applied to majority models except DC-GAN, the results in Table~\ref{tab2} show that StackGAN++ only showed slight improvement over its predecessor, StackGAN, for text-to-image synthesis. However, StackGAN++ did introduce a very worthy enhancement for unconditional image generation by organizing the generators and discriminators in a ``tree-like'' structure. This indicates that revising the structures of the discriminators and/or generators can bring a moderate level of improvement in text-to-image synthesis.}

\textcolor{black}{In addition, the results in Table~\ref{tab2} also show that DM-GAN~\citep{zhu2019dm} has the best performance, followed by Obj-GAN~\citep{DBLP:journals/corr/abs-1902-10740}. Notice that both DM-GAN and Obj-GAN are most recently developed methods in the field (both published in 2019), indicating that research in text to image synthesis is continuously improving the results for better visual perception and interception. Technical wise, DM-GAN~\citep{zhu2019dm} is a model using dynamic memory to refine fuzzy image contents initially generated from the GAN networks. A memory writing gate is used for DM-GAN to select important text information and generate images based on he selected text accordingly. On the other hand, Obj-GAN~\citep{DBLP:journals/corr/abs-1902-10740} focuses on object centered text-to-image synthesis. The proposed framework of Obj-GAN consists of a layout generation, including a bounding box generator and a shape generator, and an object-driven attentive image generator. The designs and advancement of DM-GAN and Obj-GAN indicate that research in text-to-image synthesis is advancing to put more emphasis on the image details and text semantics for better understanding and perception. }



\begin{table}
	\centering
	
	\caption{A summary of performance of different methods with respect to the three benchmark datasets and four performance metrics: Inception Score (IS), Frechet Inception Distance (FID), Human Classifier (HC), and SSIM scores. The generative adversarial networks inlcude DCGAN, GAN-INT-CLS, Dong-GAN, Paired-D-GAN, StackGAN, StackGAN++, AttnGAN, ObjGAN,HDGAN, DM-GAN, TAC-GAN, Text-SeGAN, Scene Graph GAN, and MirrorGAN. The three benchmark datasets include CUB, Oxford, and COCO datasets. A dash indicates that no data was found.}
	
	\bigskip
	
	%

	\resizebox{0.9\columnwidth}{!}{%
		\begin{tabular}{|l|l|l|l|l|l|l|c|c|c|c|c|c|}
			\hline
			& \multicolumn{12}{c|}{Datasets \& Metrics}\\
			Methods & \multicolumn{4}{c|}{CUB}  & \multicolumn{4}{c|}{COCO}  & \multicolumn{4}{c|}{Oxford} \\
			\cline{2-13}
			& IS &  FID & HC  & SSIM & IS &   FID &   HC  & SSIM & IS &  FID & HC & SSIM \\ \hline
			DCGAN   & 2.88  & 68.79  & 2.76 & - & 7.88  & 60.82  & 1.82 & -  & 2.66  & 79.55  & 1.84 & - \\
			GAN-INT-CLS  & 2.32 & 68.79 & 2.75 & - & 7.95  & 60.62  & 1.93 & -  & 2.69  & 79.55  & 1.90 & -  \\
			Dong-GAN & -  & - & - & - & -  & -  & - & -  & 4.14 & -  & - & - \\
			Paired-D-GAN    & - & - & - & - & -  & -  & - & - & 4.49  & -  & -  & -  \\
			
			StackGAN  & 3.74 & 51.89 & 1.27 & 0.234 & 8.60  & 74.05  & 1.14 & -  & 3.21  & 55.28  & 1.16 & -  \\
			StackGAN++ & 4.09  & 15.30  & 1.17 & - & 8.40  & 81.59  & 1.55 & -  & 3.27  & 48.68  & 1.27 & - \\
			
			AttnGAN    & 4.39  & 23.98 & - & - & 26.36  & 35.49  & - & - & -  & -  & -  & -  \\
			Obj-GAN  & - & - & - & - & 30.11  & 20.75  & - & -  & - & - & - & - \\
			
			HDGAN      & 4.20 & -  & - & 0.215 & 12.04  & -  & - & -  & 3.52  & -  & - & -  \\
			DM-GAN    & 4.82  & 16.09 & - & - & 31.06  & 32.64  & - & - & -  & -  & -  & - \\
			
			TAC-GAN    & - & - & - & - & - & -  & - & - & 3.50  & -  & -  & -  \\
			
			Text-SeGAN    & - & - & - & - & -  & -  & - & - & 4.1  & -  & -  & -  \\
			Scene Graph GAN      & - & -  & - & - & 7.40 & -  & - & - & - & -  & -  & -  \\
			MirrorGAN  & 4.61 & - & - & - & 26.88  & -  & - & - & - & - & - & - \\
			
			\hline
		\end{tabular}%
	}
	\bigskip
	\label{tab2}
\end{table}

\subsection{Notable Mentions}
It is worth noting that although this survey mainly focuses on text-to-image synthesis, there have been other applications of GANs in broader image synthesis field that we found fascinating and worth dedicating a small section to. For example, \citet{semilat} used Sem-Latent GANs to generate images of faces based on facial attributes, producing impressive results that, at a glance, could be mistaken for real faces. \citet{sat,wla}, and \citet{dvsa} demonstrated great success in generating text descriptions from images (image captioning) with great accuracy, with \citet{sat} using an attention-based model that automatically learns to focus on salient objects and \citet{dvsa} using deep visual-semantic alignments. Finally, there is a contribution made by StackGAN++ that was not mentioned in the dedicated section due to its relation to unconditional image generation as opposed to conditional, namely a color-regularization term \citep{stackv2}. This additional term aims to keep the samples generated from the same input at different stages more consistent in color, which resulted in significantly better results for the unconditional model.

\section{Conclusion}
The recent advancement in text-to-image synthesis research opens the door to several compelling methods and architectures. The main objective of text-to-image synthesis initially was to create images from simple labels, and this objective later scaled to natural languages. In this paper, we reviewed novel methods that generate, in our opinion, the most visually-rich and photo-realistic images, from text-based natural language. These generated images often rely on generative adversarial networks (GANs), deep convolutional decoder networks, and multimodal learning methods.

\textcolor{black}{In the paper, we first proposed a taxonomy to organize GAN based text-to-image synthesis frameworks into four major groups: semantic enhancement GANs, resolution enhancement GANs, diversity enhancement GANs, and motion enhancement GANs. The taxonomy provides a clear roadmap to show the motivations, architectures, and difference of different methods, and also outlines their evolution timeline and relationships. Following the proposed taxonomy, we reviewed important features of each method and their architectures. We indicated the model definition and key contributions from some advanced GAN framworks, including StackGAN, StackGAN++, AttnGAN, DC-GAN, AC-GAN, TAC-GAN, HDGAN, Text-SeGAn, StoryGAN \textit{etc.} Many of the solutions surveyed in this paper tackled the highly complex challenge of generating photo-realistic images beyond swatch size samples. In other words, beyond the work of \citep{reed1} in which images were generated from text in 64$\times$64 tiny swatches. Lastly, all methods were evaluated on datasets that included birds, flowers, humans, and other miscellaneous elements. We were also able to allocate some important papers that were as impressive as the papers we finally surveyed. Though, these notable papers have yet to contribute directly or indirectly to the expansion of the vast computer vision AI field. Looking into the future, an excellent extension from the works surveyed in this paper would be to give more independence to the several learning methods (\textit{e.g.} less human intervention) involved in the studies as well as increasing the size of the output images.}  \par

\section*{acknowledgements}

\section*{conflict of interest}
The authors declare that there is no conflict of interest regarding the publication of this article.


\bibliography{sample}

	\end{sloppypar}

\end{document}